%% file: template.tex
\documentclass[letterpaper,times]{IONconf}

\input{utils.tex}
\usepackage[utf8]{inputenc} 
\usepackage[T1]{fontenc}    

\title{Data-Driven Protection Levels for Camera and 3D Map-based Safe Urban Localization}

\author{
    Shubh~Gupta and Grace~Xingxin~Gao, \textit{Stanford~University}
	}

\begin{document}
\maketitle


\input{abstract.tex}
\input{introduction.tex}

\input{related_work.tex}
\input{problem_formulation.tex}

\input{types_uncertainty.tex}

\input{ddpl.tex}
\input{results.tex}

\input{conclusion.tex}

\section*{Acknowledgements}
This material is based upon work supported by the National Science Foundation under award \#2006162.

\bibliographystyle{unsrtnat}
\bibliography{references} 


\end{document}

%% file: utils.tex
\usepackage{amsmath,amssymb}
\usepackage{graphicx}

\usepackage{array}
\usepackage{caption}
\usepackage{booktabs}       
\usepackage{amsfonts}       
\usepackage{multirow}
\usepackage{subcaption}
\usepackage{floatrow}
\usepackage{xcolor}
\usepackage{bbm}
\usepackage{nicefrac}       
\usepackage{microtype}      
\usepackage{soulutf8}

\usepackage[numbers]{natbib}

\usepackage{hyperref}       
\usepackage{url}            

\graphicspath{{Figures/}}
\DeclareGraphicsExtensions{.pdf,.jpeg,.png,.pdf_tex}

\newcommand\numberthis{\addtocounter{equation}{1}\tag{\theequation}}

\usepackage{tikz}

\usepackage{algorithm, algorithmic}

\newfloatcommand{capbtabbox}{table}[][\FBwidth]

\newlength\myindent
\usepackage{eqparbox}

\newcommand{\R}{\ensuremath{\mathbb{R}}}

\newcommand{\E}{\mathbb{E}}

\renewcommand{\P}{\mathbb{P}}

\DeclareMathOperator{\median}{median}
\DeclareMathOperator{\atan2}{atan2}

\newcommand{\Prob}{\mathbb{P}}

\DeclareRobustCommand{\ghl}[1]{#1}
\DeclareRobustCommand{\chl}[1]{#1}
\newcommand{\gmhl}[1]{#1}
\newcommand{\cmhl}[1]{#1}

\def\*#1{\mathbf{#1}}
\def\$*#1{\boldsymbol{#1}}

%% file: abstract.tex
\section*{Abstract}
	Reliably assessing the error in an estimated vehicle position is integral for ensuring the vehicle's safety in urban environments. Many existing approaches use GNSS measurements to characterize protection levels (PLs) as probabilistic upper bounds on the position error. However, GNSS signals might be reflected or blocked in urban environments, and thus additional sensor modalities need to be considered to determine PLs. In this paper, we propose an approach for computing PLs by matching camera image measurements to a LiDAR-based 3D map of the environment. We specify a Gaussian mixture model probability distribution of position error using deep neural network-based data-driven models and statistical outlier weighting techniques. From the probability distribution, we compute the PLs by evaluating the position error bound using numerical line-search methods. Through experimental validation with real-world data, we demonstrate that the PLs computed from our method are reliable bounds on the position error in urban environments.

%% file: introduction.tex
\section{Introduction}

In recent years, research on autonomous navigation for urban environments has been garnering increasing attention. Many publications have targeted different aspects of navigation such as route planning~\cite{delling_customizable_2017}, perception~\cite{jensen_vision_2016} and localization~\cite{wolcott_robust_2017,caselitz_monocular_2016}. For trustworthy operation in each of these aspects, assessing the level of safety of the vehicle from potential system failures is critical. However, fewer works have examined the problem of safety quantification for autonomous vehicles.

In the context of satellite-based localization, safety is typically addressed via integrity monitoring (IM)~\cite{spilker_jr_global_1996}. Within IM, protection levels specify a statistical upper bound on the error in an estimated position of the vehicle, which can be trusted to enclose the position errors with a required probabilistic guarantee. To detect an unsafe estimated vehicle position, these protection levels are compared with the maximum allowable position error value, known as the alarm limit. Various methods~\cite{jiang_new_2016, cezon_analysis_2013, tran_kalman_2019} have been proposed over the years for computing protection levels, however, most of these approaches focus on GNSS-only navigation. These approaches do not directly apply to GNSS-denied urban environments, where visual sensors are becoming increasingly preferred~\cite{badue_self-driving_2021}. Although various options in visual sensors exist in the market, camera sensors are inexpensive, lightweight, and have been widely employed in industry. For quantifying localization safety in GNSS-denied urban environments, there is thus a need to develop new ways of computing protection levels using camera image measurements.

 
Since protection levels are bounds over the position error, computing them from camera image measurements requires a model that relates the measurements to position error in the estimate of the vehicle location. \chl{Furthermore, since the lateral, longitudinal and vertical directions are well-defined with respect to a vehicle's location on the road, the model must estimate the maximum position error in each of these directions for computing protection levels}~\cite{reid_localization_2019}. However, characterizing such a model is not straightforward. This is because the relation between a vehicle location in an environment and the corresponding camera image measurement is complex which depends on identifying and matching structural patterns in the measurements with prior known information about the environment~\cite{wolcott_robust_2017, caselitz_monocular_2016, taira_inloc_2021, kim_stereo_2018}.  

Recently, data-driven techniques based on deep neural networks (DNNs) have demonstrated state-of-the-art performance in determining the state of the camera sensor, comprising of its position and orientation, by identifying and matching patterns in images with a known map of the environment~\cite{lyrio_image-based_2015,amato_topometric_2020, cattaneo_cmrnet_2019} or an existing database of images~\cite{sarlin_coarse_2019, taira_inloc_2021}. By leveraging datasets consisting of multiple images with known camera states in an environment, these approaches train a DNN to model the relationship between an image and the corresponding state. However, the model characterized by the DNN can often be erroneous or brittle. For instance, recent research has shown that the output of a DNN can change significantly with minimal changes to the inputs~\cite{recht_imagenet_2019}. Thus, for using DNNs to determine the position error, uncertainty in the output of the DNN must also be addressed.

DNN-based algorithms consider two types of uncertainty~\cite{kendall_what_2017,loquercio_general_2020}. \ghl{\textit{Aleatoric} or statistical uncertainty results from the noise present in the inputs to the DNN, due to which a precise output cannot be produced}. For camera image inputs, sources of noise include illumination changes, occlusion or the presence of visually ambiguous structures, such as windows tessellated along a wall~\cite{kendall_what_2017}. \ghl{On the other hand, \textit{epistemic} or systematic uncertainty exists within the model itself}. Sources of epistemic uncertainty include poorly determined DNN model parameters as well as external factors that are not considered in the model~\cite{kiureghian_aleatory_2009}, such as environmental features that might be ignored by the algorithm while matching the camera images to the environment map.

While aleatoric uncertainty is typically modeled as the input-dependent variance in the output of the DNN~\cite{kendall_what_2017, mcallister_concrete_2017,yang_d3vo_2020}, epistemic uncertainty relates to the DNN model and, therefore, requires further deliberation. Existing approaches approximate epistemic uncertainty by assuming a probability distribution over the weight parameters of the DNN to represent the ignorance about the correct parameters~\cite{kendall_modelling_2016, gal_dropout_2016,blundell_weight_2015}. However, these approaches assume that a correct value of the parameters exists and that the probability distribution over the weight parameters captures the uncertainty in the model, both of which do not necessarily hold in practice~\cite{smith_understanding_2018}. This inability of existing DNN-based methods to properly characterize uncertainty limits their applicability to safety-critical applications, such as localization of autonomous vehicles. 

In this paper, we propose a novel method for computing protection levels associated with a given vehicular state estimate (position and orientation) from camera image measurements and a 3D map of the environment. This work is based on our recent ION GNSS+ 2020 conference paper~\cite{gupta_data-driven_2020} and includes additional experiments and improvements to the DNN training process. Recently, high-definition 3D environment maps in the form of LiDAR point clouds have become increasingly available through industry players such as HERE, TomTom, Waymo and NVIDIA, as well as through projects such as USGS 3DEP~\cite{lukas_3d_2016} and OpenTopography~\cite{krishnan_opentopography_2011}. Furthermore, LiDAR-based 3D maps are more robust to noise from environmental factors, such as illumination and weather, than image-based maps\cite{wang_urban_2020}. \ghl{Hence, we use LiDAR-based 3D point cloud maps in our approach.}

\begin{figure}[t!]
	\centering
	\includegraphics[width=\textwidth]{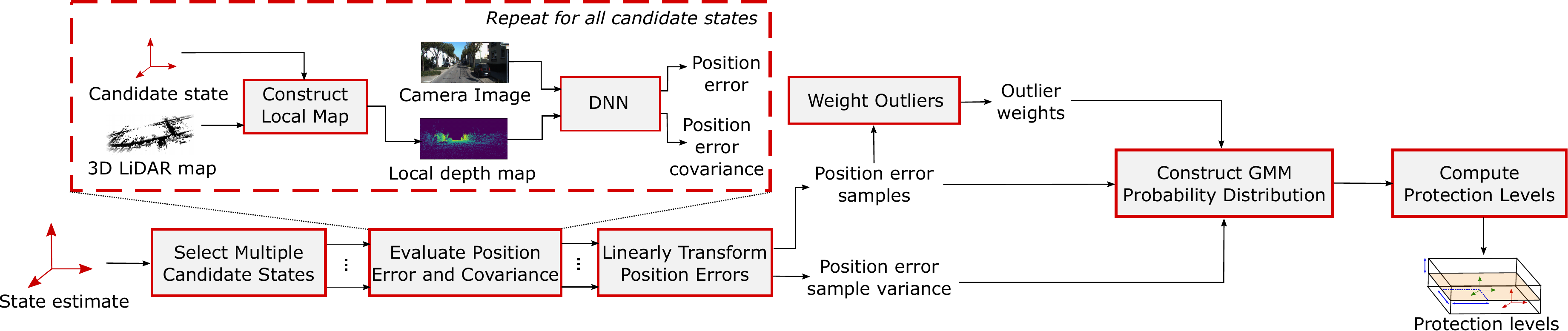}
	\caption{\ghl{Architecture of our proposed approach for computing protection levels. Given a state estimate, multiple candidate states are selected from its neighborhood and the corresponding position error and the covariance matrix for each candidate state are evaluated using the DNN. The position errors and covariance are then linearly transformed to obtain samples of the state estimate position error and variance, which are then weighted to determine outliers. Finally, the position error samples, outlier weights and variance are combined to construct a Gaussian Mixture Model probability distribution, from which the lateral, longitudinal and vertical protection levels are computed through numerical evaluation of its probability intervals.}}
	\label{fig:arch}
\end{figure}

Previously, CMRNet~\cite{cattaneo_cmrnet_2019} has been proposed as a DNN-based approach for determining the vehicular state from camera images and a LiDAR-based 3D map. \ghl{In our approach, we extend the DNN architecture proposed in}~\cite{cattaneo_cmrnet_2019} \ghl{to model the position error and the covariance matrix (aleatoric uncertainty) in the vehicular state estimate}. To assess the epistemic uncertainty in the position error, we evaluate the DNN position error outputs at multiple candidate states in the vicinity of the state estimate, and combine the outputs into samples of the state estimate position error. Fig. \ref{fig:arch} shows the architecture of our proposed approach. Given a state estimate, we first select multiple candidate states from its neighborhood. \ghl{Using the DNN, we then evaluate the position error and covariance for each candidate state by comparing the camera image measurement with a local map constructed from the candidate state and the 3D environment map. Next, we linearly transform the position error and covariance outputs from the DNN with the relative positions of candidate states into samples of the state estimate position error and variance. We then separate these samples into the lateral, longitudinal and vertical directions and weight the samples to mitigate the impact of outliers in each direction. Subsequently, we combine the position error samples, outlier weights, and variance samples to construct a Gaussian mixture model probability distribution of the position error in each direction, and numerically evaluate its intervals to compute protection levels.} 

Our main contributions are as follows:
\begin{enumerate}
	\item \ghl{We extend the CMRNet}~\cite{cattaneo_cmrnet_2019} \ghl{architecture to model both the position error in the vehicular state estimate and the associated covariance matrix. Using the 3D LiDAR-based map of the environment, we first construct a local map representation with respect to the vehicular state estimate. Then, we use the DNN to analyze the correspondence between the camera image measurement and the local map for determining the position error and the covariance matrix.}
	\item We develop a novel method for capturing epistemic uncertainty in the DNN position error output. Unlike existing approaches which assume a probability distribution over DNN weight parameters, we directly analyze different position errors that are determined by the DNN for multiple candidate states selected from within a neighborhood of the state estimate.
	The position error outputs from the DNN corresponding to the candidate states are then linearly combined with the candidate states' relative position from the state estimate, to obtain an empirical distribution of the state estimate position error.
	\item We design an outlier weighting scheme to account for possible errors in the DNN output at inputs that differ from the training data. Our approach weighs the position error samples from the empirical distribution using a robust outlier detection metric, known as robust Z-score~\cite{iglewicz_how_1993}, along the lateral, longitudinal and vertical directions individually.
	\item \ghl{We construct the lateral, longitudinal and vertical protection levels as intervals over the probability distribution of the position error. We model this probability distribution as a Gaussian Mixture Model}~\cite{lindsay_mixture_1995} \ghl{from the position error samples, DNN covariance and outlier weights.}          
	\item We demonstrate the applicability of our approach in urban environments by experimentally validating the protection levels computed from our method on real-world data with multiple camera images and different state estimates.         
\end{enumerate}


The remainder of this paper is structured as follows: Section II discusses related work. Section III formulates the problem of estimating protection levels. Section IV describes the two types of uncertainties considered in our approach. Section V details our algorithm. Section VI presents the results from experimentation with real-world data. We conclude the paper in Section VII.


%% file: related_work.tex
\section{Related Work}


Several methods have been developed over the years which characterize protection levels in the context of GNSS-based urban navigation. Jiang and Wang~\cite{jiang_new_2016} compute horizontal protection levels using an iterative search-based method and test statistic based on the bivariate normal distribution. Cezón \emph{et al.}~\cite{cezon_analysis_2013} analyze methods which utilize the isotropy of residual vectors from the least-squares position estimation to compute the protection levels. Tran and Presti~\cite{tran_kalman_2019} combine Advanced Receiver Autonomous Integrity Monitoring (ARAIM) with Kalman filtering, and compute the protection levels by considering the set of position solutions which arise after excluding faulty measurements. These approaches compute the protection levels by deriving the mathematical relation between measurement and position domain errors. However, such a relation is difficult to formulate with camera image measurements and a LiDAR-based 3D map, since the position error in this case depends on various factors such as the structure of buildings in the environment, available visual features and illumination levels.

\chl{Previous works have proposed IM approaches for LiDAR and camera-based navigation where the vehicle is localized by associating identified landmarks with a stored map or a database. Joerger \emph{et al.}}~\cite{joerger_quantifying_2019}\chl{ developed a method to quantify integrity risk for LiDAR-based navigation algorithms by analyzing failures of feature extraction and data association subroutines. Zhu \emph{et al.}}~\cite{zhu_quantifying_2020} \chl{derived a bound on the integrity risk in camera-based navigation using EKF caused by incorrect feature associations. However, these IM approaches have been developed for localization algorithms based on data-association and cannot be directly applied to many recent camera and LiDAR-based localization techniques which use deep learning to model the complex relation between measurements and the stored map or the database. Furthermore, these IM techniques do not estimate protection levels, which is the focus of our work.}

Deep learning has been widely applied for determining position information from camera images.  Kendall \emph{et al.}~\cite{kendall_posenet_2015} train a DNN using images from a single environment to learn a relation between the image and the camera 6-DOF pose. Taira \emph{et al.}~\cite{taira_inloc_2021} learn image features using a DNN and apply feature extraction and matching techniques to estimate the 6-DOF camera pose relative to a known 3D map of the environment. Sarlin \emph{et al.}~\cite{sarlin_coarse_2019} develop a deep learning-based 2D-3D matching technique to obtain 6-DOF camera pose from images and a 3D environment model. However, these approaches do not model the corresponding uncertainty associated with the estimated camera pose, or account for failures in DNN approximation~\cite{smith_understanding_2018}, which is necessary for characterizing safety measures such as protection levels. 


Some recent works have proposed to estimate the uncertainty associated with deep learning algorithms.  Kendall and Cipolla~\cite{kendall_modelling_2016} estimate the uncertainty in DNN-based camera pose estimation from images, by evaluating the network multiple times through dropout~\cite{gal_dropout_2016}. Loquercio \emph{et al.}~\cite{loquercio_general_2020} propose a general framework for estimating uncertainty in deep learning as variance computed from both aleatoric and epistemic sources. McAllister \emph{et al.}~\cite{mcallister_concrete_2017} suggest using Bayesian deep learning to determine uncertainty and quantify safety in autonomous vehicles, by placing probability distributions over DNN weights to represent the uncertainty in the DNN model. Yang \emph{et al.}~\cite{yang_d3vo_2020} jointly estimate the vehicle odometry, scene depth and uncertainty from sequential camera images. However, the uncertainty estimates from these algorithms do not take into account the inaccuracy of the trained DNN model, or the influence of the underlying environment structure on the DNN outputs. In our approach, we evaluate the DNN position error outputs at inputs corresponding to multiple states in the environment, and utilize these position errors for characterizing uncertainty both from inaccuracy in the DNN model as well as from the environment structure around the state estimate.     

To the best of our knowledge, our approach is the first that applies data-driven algorithms for computing protection levels by characterizing the uncertainty from different error sources. The proposed method seeks to leverage the high-fidelity function modeling capability of DNNs and combine it with techniques from robust statistics and integrity monitoring to compute robust protection levels using camera image measurements and 3D map of the environment. 

%% file: problem_formulation.tex
\section{Problem Formulation}
We consider the scenario of a vehicle navigating in an urban environment using measurements acquired by an on-board camera. The 3D LiDAR map of the environment $\mathcal{M}$ that consists of points $\*p \in \R^3$ is assumed to be pre-known from either openly available repositories~\cite{lukas_3d_2016, krishnan_opentopography_2011} or from Simultaneous Localization and Mapping algorithms~\cite{cadena_past_2016}. 

\ghl{The vehicular state $\*s_t = [\*x_t, \*o_t]$ at time $t$ is a 7-element vector comprising of its 3D position $\*x_t = [x_t, y_t, z_t]^\top \in \R^3$ along $x, y$ and $z$-dimensions and 3D orientation unit quaternion $\*o_t = [o_{1, t}, o_{2, t}, o_{3, t}, o_{4, t}] \in \textrm{SU}(2)$.} The vehicle state estimates over time are denoted as $\{\*s_t\}_{t=1}^{T_{\text{max}}}$ where $T_{\text{max}}$ denotes the total time in a navigation sequence. At each time $t$, the vehicle captures an RGB camera image $I_t \in \R^{l \times w \times 3}$ from the on-board camera, where $l$ and $w$ denote pixels along length and width dimensions, respectively. 


\ghl{Given an integrity risk specification $IR$, our objective is to compute the lateral protection level $PL_{lat, t}$, longitudinal protection level $PL_{lon, t}$, and vertical protection level $PL_{vert, t}$ at time $t$, which denote the maximal bounds on the position error magnitude with a probabilistic guarantee of at least $1-IR$. Considering $x, y$ and $z$-dimensions in the rotational frame of the vehicle} 
\begin{align*}
	\gmhl{PL_{lat, t}} &\gmhl{= \sup\left\{\rho \mid \Prob \left(|x_t-x^*_t| \le \rho \right) \le 1 - IR \right\} }\\
    \gmhl{PL_{lon, t}} &\gmhl{= \sup\left\{\rho \mid \Prob \left(|y_t-y^*_t| \le \rho \right) \le 1 - IR \right\} }\\
    PL_{vert, t} &= \sup\left\{\rho \mid \Prob \left(|z_t-z^*_t| \le \rho \right) \le 1 - IR \right\}, \numberthis
    \label{eqn:pl}
\end{align*}
where $\*x^*_t = [x^*_t, y^*_t, z^*_t]$ denotes the unknown true vehicle position at time $t$.

%% file: types_uncertainty.tex
\section{Types of Uncertainty in Position Error}

Protection levels for a state estimate $\*s_t$ at time $t$ depend on the uncertainty in determining the associated position error $\Delta \*x_t = [\Delta x_t, \Delta y_t, \Delta z_t]$ between the state estimate position $\*x_t$ and the true position $\*x^*_t$ from the camera image $I_t$ and the environment map $\mathcal{M}$. We consider two different kinds of uncertainty, which are categorized by the source of inaccuracy in determining the position error $\Delta \*x_t$: aleatoric and epistemic uncertainty.

\subsection{Aleatoric Uncertainty}
Aleatoric uncertainty refers to the uncertainty from noise present in the camera image measurements $I_t$ and the environment map $\mathcal{M}$, due to which a precise value of the position error $\Delta \*x_t$ cannot be determined. \ghl{Existing DNN-based localization approaches model the aleatoric uncertainty as a covariance matrix with only diagonal entries}~\cite{kendall_what_2017, mcallister_concrete_2017,yang_d3vo_2020} \ghl{or with both diagonal and off-diagonal terms}~\cite{russell_multivariate_2019, liu_deep_2018}. \ghl{Similar to the existing approaches, we characterize the aleatoric uncertainty by using a DNN to model the covariance matrix $\Sigma_t$ associated with the position error $\Delta \*x_t$. We consider both nonzero diagonal and off-diagonal terms in $\Sigma_t$ to model the correlation between $x, y$ and $z$-dimension uncertainties, such as along the ground plane. } 

Aleatoric uncertainty by itself does not accurately represent the uncertainty in determining the position error. This is because aleatoric uncertainty assumes that the noise present in training data also represents the noise in all future inputs and that the DNN approximation is error-free. These assumptions fail in scenarios when the input at evaluation time is different from the training data or when the input contains features that occur rarely in the real world~\cite{smith_understanding_2018}. Thus, relying purely on aleatoric uncertainty can lead to an overconfident estimates of the position error uncertainty~\cite{kendall_what_2017}.

\begin{figure}[t!]
	\centering
	\hspace{3.5cm} \def\svgwidth{0.5\linewidth} 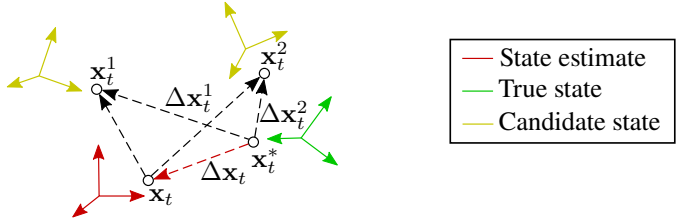
	\caption{Position error $\Delta \*x_t$ in the state estimate position $\*x_t$ is a linear combination of the position error $\Delta \*x^i_t$ in position $\*x^i_t$ of any candidate state $s^i_t$ and the relative position vector between $\*x^i_t$ and $\*x_t$.}
	\label{fig:epistemic}
\end{figure}

\subsection{Epistemic Uncertainty}
Epistemic uncertainty relates to the inaccuracies in the model for determining the position error $\Delta \*x_t$. In our approach, we characterize the epistemic uncertainty by leveraging a geometrical property of the position error $\Delta \*x_t$, where for the same camera image $I_t$, $\Delta \*x_t$ can be obtained by linearly combining the position error $\Delta \*x'_t$ computed for any \emph{candidate state} $\*s'_t$ and the relative position of $\*s'_t$ from the state estimate $\*s_t$ (Fig. \ref{fig:epistemic}). Hence, using known relative positions and orientations of $N_C$ candidate states $\{\*s_t^1, \ldots, \*s_t^{N_C}\}$ from $\*s_t$, we transform the different position errors $\{\Delta \*x_t^1, \ldots, \Delta \*x_t^{N_C}\}$ determined for the candidate states into samples of the state estimate position error $\Delta \*x_t$. The empirical distribution comprised of these position error samples characterizes the epistemic uncertainty in the position error estimated using the DNN.

%% file: Figures/epistemic.pdf_tex
\begingroup%
  \makeatletter%
  \providecommand\color[2][]{%
    \errmessage{(Inkscape) Color is used for the text in Inkscape, but the package 'color.sty' is not loaded}%
    \renewcommand\color[2][]{}%
  }%
  \providecommand\transparent[1]{%
    \errmessage{(Inkscape) Transparency is used (non-zero) for the text in Inkscape, but the package 'transparent.sty' is not loaded}%
    \renewcommand\transparent[1]{}%
  }%
  \providecommand\rotatebox[2]{#2}%
  \newcommand*\fsize{\dimexpr\f@size pt\relax}%
  \newcommand*\lineheight[1]{\fontsize{\fsize}{#1\fsize}\selectfont}%
  \ifx\svgwidth\undefined%
    \setlength{\unitlength}{402.48942126bp}%
    \ifx\svgscale\undefined%
      \relax%
    \else%
      \setlength{\unitlength}{\unitlength * \real{\svgscale}}%
    \fi%
  \else%
    \setlength{\unitlength}{\svgwidth}%
  \fi%
  \global\let\svgwidth\undefined%
  \global\let\svgscale\undefined%
  \makeatother%
  \begin{picture}(1,0.32245686)%
    \lineheight{1}%
    \setlength\tabcolsep{0pt}%
    \put(0,0){\includegraphics[width=\unitlength,page=1]{epistemic.pdf}}%
    \put(0.21207036,0.02470838){\color[rgb]{0,0,0}\makebox(0,0)[lt]{\lineheight{1.25}\smash{\begin{tabular}[t]{l}$\mathbf{x}_t$\end{tabular}}}}%
    \put(0.1244938,0.20768293){\color[rgb]{0,0,0}\makebox(0,0)[lt]{\lineheight{1.25}\smash{\begin{tabular}[t]{l}$\mathbf{x}^1_t$\end{tabular}}}}%
    \put(0.3799137,0.22990716){\color[rgb]{0,0,0}\makebox(0,0)[lt]{\lineheight{1.25}\smash{\begin{tabular}[t]{l}$\mathbf{x}^2_t$\end{tabular}}}}%
    \put(0.23593558,0.16721507){\color[rgb]{0,0,0}\makebox(0,0)[lt]{\lineheight{1.25}\smash{\begin{tabular}[t]{l}$\Delta \mathbf{x}^1_t$\end{tabular}}}}%
    \put(0.28907705,0.05555183){\color[rgb]{0,0,0}\makebox(0,0)[lt]{\lineheight{1.25}\smash{\begin{tabular}[t]{l}$\Delta \mathbf{x}_t$\end{tabular}}}}%
    \put(0.3767168,0.1422201){\color[rgb]{0,0,0}\makebox(0,0)[lt]{\lineheight{1.25}\smash{\begin{tabular}[t]{l}$\Delta \mathbf{x}^2_t$\end{tabular}}}}%
    \put(0.36558037,0.07593567){\color[rgb]{0,0,0}\makebox(0,0)[lt]{\lineheight{1.25}\smash{\begin{tabular}[t]{l}$\mathbf{x}^*_t$\end{tabular}}}}%
    \put(0,0){\includegraphics[width=\unitlength,page=2]{epistemic.pdf}}%
    \put(0.73691663,0.22707346){\color[rgb]{0,0,0}\makebox(0,0)[lt]{\lineheight{1.25}\smash{\begin{tabular}[t]{l}State estimate\end{tabular}}}}%
    \put(0.73499168,0.18052648){\color[rgb]{0,0,0}\makebox(0,0)[lt]{\lineheight{1.25}\smash{\begin{tabular}[t]{l}True state\end{tabular}}}}%
    \put(0.73553625,0.13040792){\color[rgb]{0,0,0}\makebox(0,0)[lt]{\lineheight{1.25}\smash{\begin{tabular}[t]{l}Candidate state\end{tabular}}}}%
  \end{picture}%
\endgroup%

%% file: ddpl.tex
\section{Data-Driven Protection Levels}
\label{sec:ddpl}
This section details our algorithm for computing data-driven protection levels for the state estimate $\*s_t$ at time $t$, using the camera image $I_t$ and environment map $\mathcal{M}$. First, we describe the method for generating local representations of the 3D environment map $\mathcal{M}$ with respect to the state estimate $\*s_t$. Then, we illustrate the architecture of the DNN. Next, we discuss the loss functions used in DNN training. We then detail the method for selecting multiple candidate states from the neighborhood of the state estimate $\*s_t$. \ghl{Using position errors and covariance matrix evaluated from the DNN for each of these candidate states, we then illustrate the process for transforming the candidate state position errors into multiple samples of the state estimate position error. Then, to mitigate the impact of outliers in the computed position error samples in each of the lateral, longitudinal and vertical directions, we detail the procedure for computing outlier weights. Next, we characterize the probability distribution over position error in lateral, longitudinal and vertical directions. Finally, we detail the approach for determining protection levels from the probability distribution by numerical methods.}

\input{local_map.tex}

\input{dnn_arch.tex}

\input{losses.tex}


\input{cand_state.tex}

\input{multi_samp.tex}

\input{outlier_wt.tex}


\input{prob_dist.tex}

\input{pl_comp.tex}

%% file: local_map.tex
\subsection{Local Map Construction}
\label{sec:localmap}
A local representation of the 3D LiDAR map of the environment captures the environment information in the vicinity of the state estimate $\*s_t$ at time $t$. By comparing the environment information captured in the local map with the camera image $I_t \in \R^{l \times w \times 3}$ using a DNN, we estimate the position error $\Delta \*x_t$ and covariance $\Sigma_t$ in the state estimate $\*s_t$. For computing the local maps, we utilize the LiDAR-image generation procedure described in~\cite{cattaneo_cmrnet_2019}. Similar to their approach, we generate the local map $L(\*s, \mathcal{M}) \in \R^{l \times w}$ associated with a vehicle state $\*s$ and LiDAR environment map $\mathcal{M}$ in two steps.
\begin{enumerate}
	\item First, we determine the rigid-body transformation matrix $H_{\*s}$ in the special Euclidean group $\textrm{SE}(3)$ corresponding to the vehicle state $\*s$
	\begin{equation}
		H_{\*s} = \left[\begin{matrix}
			R_{\*s} & T_{\*s} \\
			\*0_{1\times 3} & 1
		\end{matrix}\right] \in \textrm{SE}(3),
	\end{equation}       
	where
	\begin{enumerate}
		\item[--] \ghl{$R_{\*s}$ denotes the rotation matrix corresponding to the orientation quaternion elements $\*o = [o_1, o_2, o_3, o_4]$ in the state $\*s$} 
		\item[--] $T_{\*s}$ denotes the translation vector corresponding to the position elements $\*x = [x, y, z]$ in the state $\*s$.    
	\end{enumerate}
	Using the matrix $H_{\*s}$, we rotate and translate the points in the map $\mathcal{M}$ to the map $\mathcal{M}_{\*s}$ in the reference frame of the state $\*s$
	\begin{equation}
		\mathcal{M}_{\*s} = \{\left[\begin{matrix}I_{3\times 3} & \*0_{3\times 1}\end{matrix}\right] \cdot H_{\*s} \cdot \left[\begin{matrix}\*p^\top & 1\end{matrix}\right]^\top \mid \*p \in \mathcal{M}\},
	\end{equation}
	where $I$ denotes the identity matrix.
	
	For maintaining computational efficiency in the case of large maps, we use the points in the LiDAR map $\mathcal{M}_\*s$ that lie in a sub-region around the state $\*s$ and in the direction of the vehicle orientation.
	
	\item In the second step, we apply the occlusion estimation filter presented in~\cite{pintus_real-time_2011} to identify and remove occluded points along rays from the camera center. For each pair of points $(\*p^{(i)}, \*p^{(j)})$ where $\*p^{(i)}$ is closer to the state $\*s$, $\*p^{(j)}$ is marked occluded if the angle between the ray from $\*p^{(j)}$ to the camera center and the line from $\*p^{(j)}$ to $\*p^{(i)}$ is less than a threshold. Then, the remaining points are projected to the camera image frame using the camera projection matrix $K$ to generate the local depth map $L(\*s, \mathcal{M})$. The $i$th point $\*p^{(i)}$ in $\mathcal{M}_\*s$ is projected as 
	\begin{align*}
		[\begin{matrix}
			p_x & p_y & c
		\end{matrix}]^\top &= K \cdot \*p^{(i)} \\
		[L(\*s, \mathcal{M})]_{(\lceil p_x/c \rceil, \lceil p_y/c \rceil)} &=  [\begin{matrix}
			0 & 0 & 1
		\end{matrix}] \cdot  \*p^{(i)}, \numberthis
	\end{align*}
	where 
	\begin{enumerate}
		\item[--] $p_x,p_y$ denote the projected 2D coordinates with scaling term $c$ 
		\item[--] $[L(\*s, \mathcal{M})]_{(p_x, p_y)}$ denotes the $(p_x,p_y)$ pixel position in the local map $L(\*s, \mathcal{M})$.     
	\end{enumerate}

\end{enumerate}
The local depth map $L(\*s, \mathcal{M})$ for state $\*s$ visualizes the environment features that are expected to be captured in a camera image obtained from the state $\*s$. However, the obtained camera image $I_t$ is associated with the true state $\*s^*_t$ that might be different from the state estimate $\*s_t$. Nevertheless, for reasonably small position and orientation differences between the state estimate $\*s_t$ and true state $\*s^*_t$, the local map $L(\*s, \mathcal{M})$ contains features that correspond with some of the features in the camera image $I_t$ that we use to estimate the position error.

\begin{figure}[t!]
	\centering
	\includegraphics[width=\textwidth]{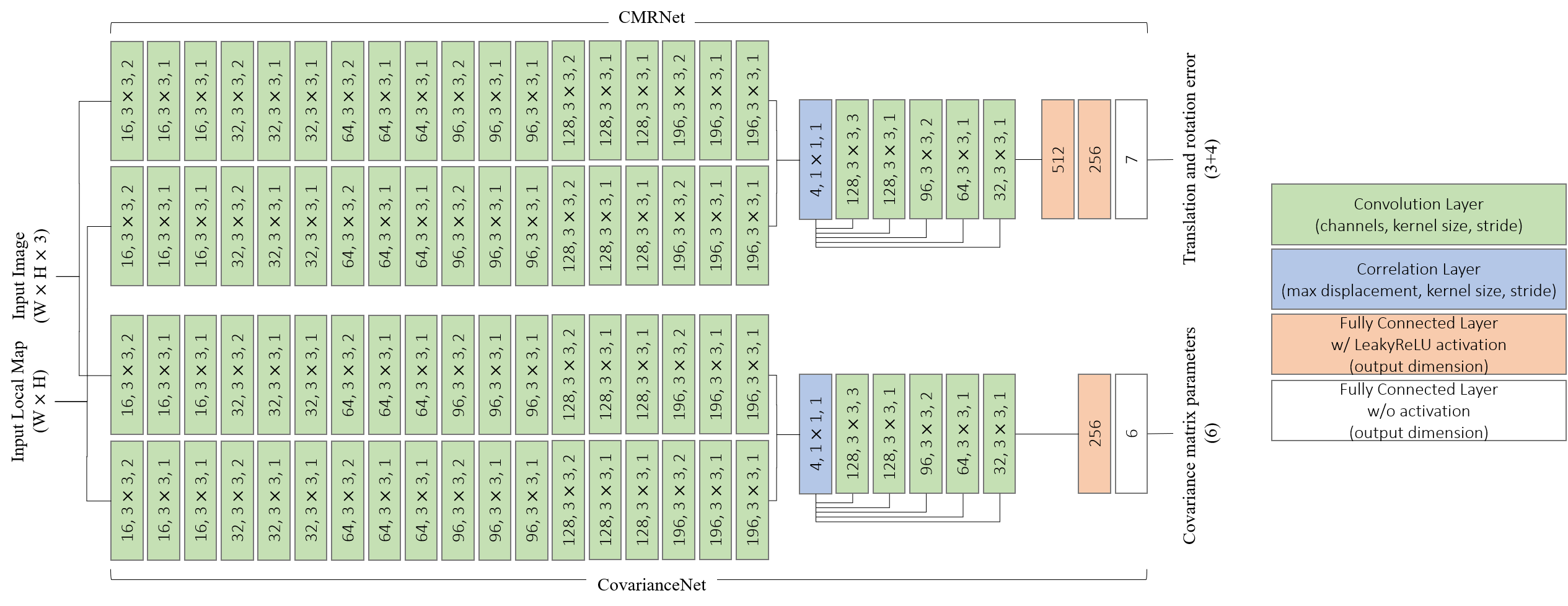}
	\caption{\ghl{Architecture of our deep neural network for estimating translation and rotation errors as well as parameters of the covariance matrix. The translation and rotation errors are determined using CMRNet}~\cite{cattaneo_cmrnet_2019}\ghl{, and employs correlation layers}~\cite{dosovitskiy_flownet_2015}\ghl{ for comparing feature representations of the camera image and the local depth map. Using a similar architecture, we design CovarianceNet which produces parameters of the covariance matrix associated with the translation error output.}}
	\label{fig:dnn_arch}
\end{figure}

%% file: dnn_arch.tex
\subsection{DNN Architecture}
\label{sec:dnn}
\ghl{We use a DNN to estimate the position error $\Delta \*x_t$ and associated covariance matrix $\Sigma_t$ by implicitly identifying and comparing the positions of corresponding features in camera image $I_t$ and the local depth map $L(\*s_t, \mathcal{M})$ associated with the state estimate $\*s_t$.} 

The architecture of our DNN is given in Fig.~\ref{fig:dnn_arch}. \ghl{Our DNN comprises of two separate modules, one for estimating the position error $\Delta \*x_t$ and other for the parameters of the covariance matrix $\Sigma_t$.} The first module for estimating the position error $\Delta \*x_t$ is based on CMRNet~\cite{cattaneo_cmrnet_2019}. CMRNet was originally proposed as an algorithm to iteratively determine the position and orientation of a vehicle using a camera image and 3D LiDAR map, starting from a provided initial state. \chl{For determining position error $\Delta \*x_t$ using CMRNet, we use the state estimate $\*s_t$ as the provided initial state and the corresponding DNN translation $\Delta \tilde{\*x}_t$ and rotation $\Delta \tilde{\*r}$ error output for transforming the state $\*s_t$ towards the true state $\*s^*_t$.} Formally, given any state $\*s$ and camera image $I_t$ at time $t$, the translation error $\Delta \tilde{\*x}$ and rotation error $\Delta \tilde{\*r}$ are expressed as
\begin{equation}
	\Delta \tilde{\*x}, \Delta \tilde{\*r} = \textrm{CMRNet}(I_t, L(\*s, \mathcal{M})).
\end{equation}
\ghl{CMRNet estimates the rotation error $\Delta \tilde{\*r}$ as a unit quaternion. Furthermore, the architecture determines both the translation error $\Delta \tilde{\*x}$ and rotation error $\Delta \tilde{\*r}$ in the reference frame of the state $\*s$. Since the protection levels depend on the position error $\Delta \*x$ in the reference frame from which the camera image $I_t$ is captured (the vehicle reference frame), we transform the translation error $\Delta \tilde{\*x}$ to the vehicle reference frame by rotating it with the inverse of $\Delta \tilde{\*r}$}
\begin{equation}
	\cmhl{\Delta \*x = -\tilde{R}^\top \cdot \Delta \tilde{\*x},}
\end{equation}
\ghl{where $\tilde{R}$ is the $3 \times 3$ rotation matrix corresponding to the rotation error quaternion $\Delta \tilde{\*r}$.}

\chl{In the second module, we determine the covariance matrix $\Sigma$ associated with $\Delta \*x$ by first estimating the covariance matrix $\tilde{\Sigma}$ associated with the translation error $\Delta \tilde{\*x}$ obtained from CMRNet and then transforming it to the vehicle reference frame using $\Delta \tilde{\*r}$.}

\chl{We model the covariance matrix $\tilde{\Sigma}$ by following a similar approach to}~\cite{russell_multivariate_2019}. \chl{Since the covariance matrix is both symmetric and positive-definite, we consider the decomposition of $\tilde{\Sigma}$ into diagonal standard deviations $\boldsymbol{\sigma} = [\sigma_1, \sigma_2, \sigma_3]$ and correlation coefficients $\boldsymbol{\eta} = [\eta_{21}, \eta_{31}, \eta_{32}]$}
\begin{align*}
	\cmhl{[\tilde{\Sigma}]_{ii}} &\cmhl{= \sigma_i^2}\\
	\cmhl{[\tilde{\Sigma}]_{ij}} &\cmhl{= [\Sigma]_{ji} = \eta_{ij}\sigma_i\sigma_j}, \numberthis 
\end{align*} 
\chl{where $i,j \in \{1,2,3\}$ and $j<i$. We estimate these terms using our second DNN module (referred to as CovarianceNet) which has a similar network structure as CMRNet, but with $256$ and $6$ artificial neurons in the last two fully connected layers to prevent overfitting. For stable training, CovarianceNet produces logarithm of the standard deviation output, which is converted to the standard deviation by then taking the exponent. Additionally, we use \texttt{tanh} function to scale the correlation coefficient outputs $\boldsymbol{\eta}$ in CovarianceNet between $\pm 1$. Formally, given a vehicle state $\*s$ and camera image $I_t$ at time $t$, the standard deviation $\boldsymbol{\sigma}$ and correlation coefficients $\boldsymbol{\eta}$ is approximated as}
\begin{equation}
	\cmhl{\boldsymbol{\sigma}, \boldsymbol{\eta} = \textrm{CovarianceNet}(I_t, L(\*s, \mathcal{M})).}
\end{equation}
\chl{Using the constructed $\tilde{\Sigma}$ from the obtained $\boldsymbol{\sigma}, \boldsymbol{\eta}$, we obtain the covariance matrix $\Sigma$ associated with $\Delta \*x$ as}
\begin{equation}
	\cmhl{\Sigma = \tilde{R}^\top \cdot \tilde{\Sigma} \cdot \tilde{R}}.
\end{equation}
\chl{We keep the aleatoric uncertainty restricted to position domain errors in this work for simplicity, and thus treat $\Delta \tilde{\*r}$ as a point estimate. The impact of errors in estimating $\Delta \tilde{\*r}$ on protection levels is taken into consideration as epistemic uncertainty, and is discussed in more detail in Section}~\hyperref[sec:pos_samp]{V.5} and \hyperref[sec:prob_dist]{V.7}.

The feature extraction modules in CovarianceNet and CMRNet are separate since the two tasks are complementary: for estimating position error, the DNN must learn features that are robust to noise in the inputs while the variance in the estimated position error depends on the noise itself.

%% file: losses.tex
\subsection{Loss Functions}
The loss function for training the DNN must penalize position error outputs that differ from the corresponding ground truth present in the dataset, as well as penalize covariance that overestimates or underestimates the uncertainty in the position error predictions. Furthermore, the loss muss incentivize the DNN to extract useful features from the camera image and local map inputs for predicting the position error. Hence, we consider three additive components in our loss function $\mathcal{L}(\cdot)$
\begin{equation}
	\mathcal{L} = \alpha_{\textrm{Huber}}\mathcal{L}_{\textrm{Huber}}(\Delta \tilde{\*x}^*, \Delta \tilde{\*x}) + \gmhl{\alpha_{\textrm{MLE}}\mathcal{L}_{\textrm{MLE}}(\Delta \tilde{\*x}^*, \Delta \tilde{\*x}, \tilde{\Sigma})} + \alpha_{\textrm{Ang}} \mathcal{L}_{\textrm{Ang}}(\Delta \tilde{\*r}^*, \Delta \tilde{\*r}), 
\end{equation} 
where
\begin{enumerate}
	\item[--] $\Delta \tilde{\*x}^*, \Delta \tilde{\*r}^*$ denotes the vector-valued translation and rotation error in the reference frame of the state estimate $\*s$ to the unknown true state $\*s^*$
	\item[--] $\mathcal{L}_{\textrm{Huber}}(\cdot)$ denotes the Huber loss function~\cite{kotz_robust_1992} 
	\item[--] \ghl{$\mathcal{L}_{\textrm{MLE}}(\cdot)$ denotes the loss function for the maximum likelihood estimation of position error $\Delta \*x$ and covariance $\tilde{\Sigma}$}
	\item[--] $\mathcal{L}_{\textrm{Ang}}(\cdot)$ denotes the quaternion angular distance from~\cite{cattaneo_cmrnet_2019}
	\item[--] $\alpha_{\textrm{Huber}}, \alpha_{\textrm{MLE}}, \alpha_{\textrm{Ang}}$ are coefficients for weighting each loss term.       
\end{enumerate}
We employ the Huber loss $\mathcal{L}_{\textrm{Huber}}(\cdot)$ and quaternion angular distance $\mathcal{L}_{\textrm{Ang}}(\cdot)$ terms from~\cite{cattaneo_cmrnet_2019}. The Huber loss term $\mathcal{L}_{\textrm{Huber}}(\cdot)$ penalizes the translation error output $\Delta \tilde{\*x}$ of the DNN
\begin{align*}
	\mathcal{L}_{\textrm{Huber}}(\Delta \tilde{\*x}^*, \Delta \tilde{\*x}) &= \sum_{X=x,y,z} D_{\textrm{Huber}}(\Delta \tilde{X}^*, \Delta \tilde{X})\\
	D_{\textrm{Huber}}(a^*, a) &= \begin{cases}
		\frac{1}{2}(a-a^*)^2 & \textrm{for } |a-a^*| \le \delta \\
		\delta \cdot (|a-a^*| - \frac{1}{2}\delta) & \textrm{otherwise}
	\end{cases}, \numberthis 
\end{align*}
where $\delta$ is a hyperparameter for adjusting the penalty assignment to small error values.
In this paper, we set $\delta=1$. Unlike the more common mean squared error, the penalty assigned to higher error values is linear in Huber loss instead of quadratic. Thus, Huber loss is more robust to outliers and leads to more stable training as compared with squared error. The quaternion angular distance term $\mathcal{L}_{\textrm{Ang}}(\cdot)$ penalizes the rotation error output $\Delta \tilde{\*r}$ from CMRNet
\begin{align*}
	\mathcal{L}_{\textrm{Ang}}(\Delta \tilde{\*r}^*, \Delta \tilde{\*r}) &= D_{\textrm{Ang}}(\Delta \tilde{\*r}^* \times \Delta \tilde{\*r}^{-1}) \\
	D_{\textrm{Ang}}(\*q) &= \atan2\left(\sqrt{q_2^2+q_3^2+q_4^2}, |q_1|\right), \numberthis  
\end{align*}
where 
\begin{enumerate}
	\item[--] $q_i$ denotes the $i$th element in quaternion $\*q$
	\item[--] $\Delta \*r^{-1}$ denotes the inverse of the quaternion $\Delta \*r$
	\item[--] $\*q \times \*r$ here denotes element-wise multiplication of the quaternions $\*q$ and $\*r$
	\item[--] $\atan2(\cdot)$ is the two-argument version of the arctangent function.   
\end{enumerate}
Including the quaternion angular distance term $\mathcal{L}_{\textrm{Ang}}(\cdot)$ in the loss function incentivizes the DNN to learn features that are relevant to the geometry between the camera image and the local depth map. Hence, it provides additional supervision to the DNN training as a multi-task objective~\cite{zeng_deep_2015}, and is important for the stability and speed of the training process.

\chl{The maximum likelihood loss term $\mathcal{L}_{\textrm{MLE}}(\cdot)$ depends on both the translation error $\Delta \tilde{\*x}$ and covariance matrix $\tilde{\Sigma}$ estimated from the DNN. The loss function is analogous to the negative log-likelihood of the Gaussian distribution}
\begin{equation}
	\cmhl{\mathcal{L}_{\textrm{MLE}}( \Delta \tilde{\*x}^*, \Delta \tilde{\*x}, \tilde{\Sigma}) = \frac{1}{2}\log |\tilde{\Sigma}| + \frac{1}{2}(\Delta \tilde{\*x}^* - \Delta \tilde{\*x})^\top \cdot \tilde{\Sigma}^{-1} \cdot (\Delta \tilde{\*x}^* - \Delta \tilde{\*x})}
\end{equation}  
If the covariance output from the DNN has small values, the corresponding translation error is penalized much more than the translation error corresponding to a large valued covariance. Hence, the maximum likelihood loss term $\mathcal{L}_{\textrm{MLE}}(\cdot)$ incentivizes the DNN to output small covariance only when the corresponding translation error output has high confidence, and otherwise output large covariance.   

%% file: cand_state.tex
\subsection{Multiple Candidate State Selection}

To assess the uncertainty in the DNN-based position error estimation process as well as the uncertainty from environmental factors, we evaluate the DNN output at $N_C$ candidate states $\{\*s^{1}_t \ldots, \*s^{N_C}_t\}$ in the neighborhood of the state estimate $\*s_t$.

For selecting the candidate states $\{\*s^{1}_t \ldots, \*s^{N_C}_t\}$, we randomly generate multiple values of translation offset $\{\*t^1, \ldots, \*t^{N_C}\}$ and rotation offset $\{\*r^1, \ldots, \*r^{N_C}\}$ about the state estimate $\*s_t$, where $N_C$ is the total number of selected candidate states. The $i$th translation offset $\*t^i \in \R^3$ denotes translation in $x, y$ and $z$ dimensions and is sampled from a uniform probability distribution between a specified range $\pm t_{max}$ in each dimension. \ghl{Similarly, the $i$th rotation offset $\*r^i \in \textrm{SU}(2)$ is obtained by uniformly sampling between $\pm r_{max}$ angular deviations about each axis and converting the resulting rotation to a quaternion.} The $i$th candidate state $\*s^i_t$ is generated by rotating and translating the state estimate $\*s_t$ by $\*r^i$ and $\*t^i$,  respectively. Corresponding to each candidate state $\*s^i_t$, we generate a local depth map $L(\*s^i_t, \mathcal{M})$ using the procedure laid out in Section \hyperref[sec:localmap]{V.1}.

%% file: multi_samp.tex
\subsection{Linear Transformation of Position Errors}
\label{sec:pos_samp}
Using each local depth map $L(\*s^i_t, \mathcal{M})$ and camera image $I_t$ for the $i$th candidate state $\*s^i_t$ as inputs to the DNN in Section \hyperref[sec:dnn]{V.2}, we evaluate the candidate state position error $\Delta \*x^i_t$ and covariance matrix $\Sigma^i_t$. \ghl{From the known translation offset $\*t^i$ between the candidate state $\*s^i_t$ and the state estimate $\*s_t$ and the DNN-based rotation error $\Delta \tilde{\*r}_t$ in $\*s_t$, we compute the transformation matrix $H_{\*s^i_t \to \*s_t}$ for converting the candidate state position error $\Delta \*x^i_t$ to the state estimate position error $\Delta \*x_t$ in the vehicle reference frame}
\begin{equation}
	\gmhl{H_{\*s^i_t \to \*s_t} = \left[\begin{matrix}
		I_{3 \times 3} & -\tilde{R}_t^\top\*t^i
	\end{matrix}\right],}
\end{equation}
\ghl{where $I_{3 \times 3}$ denotes the identity matrix and $\tilde{R}_t$ is the $3 \times 3$ rotation matrix computed from the DNN-based rotation error $\Delta \tilde{\*r}_t$ between the state estimate $\*s_t$ and the unknown true state $\*s^*_t$. Note that the rotation offset $\*r^i$ is not used in the transformation, since we are only concerned with the position errors from the true state $\*s^*_t$ to the state estimate $\*s_t$, which are invariant to the orientation of the candidate state $\*s^i_t$.} Using the transformation matrix $H_{\*s^i_t \to \*s_t}$, we obtain the $i$th sample of the state estimate position error $\Delta \*x_t^{(i)}$
\begin{equation}
	\gmhl{\Delta \*x_t^{(i)} = H_{\*s^i_t \to \*s_t} \cdot [\begin{matrix}
		\Delta \*x^i_t & 1
	\end{matrix}]^\top = \Delta \*x^i_t - \tilde{R}_t^\top\*t^i.}
	\label{eqn:transform}
\end{equation}
\chl{We use parentheses in the notation $\Delta \*x_t^{(i)}$ for the transformed samples of the position error between the true state $\*s^*_t$ and the state estimate $\*s_t$ to differentiate from the position error $\Delta \*x^i_t$ between $\*s^*_t$ and the candidate state $\*s^i_t$. Next, we modify the candidate state covariance matrix $\Sigma^i_t$ to account for uncertainty in DNN-based rotation error $\Delta \tilde{\*r}_t$.} \chl{The resulting covariance matrix $\Sigma^{(i)}_t$ in terms of the covariance matrix $\Sigma^i_t$ for $\Delta \*x^i_t$, $\tilde{R}_t$ and $\*t^i$ is}
 \begin{equation}
    \cmhl{\Sigma^{(i)}_t = \Sigma^i_t + \textrm{Var}[\tilde{R}_t^\top\*t^i].}
 \end{equation}

 \chl{Assuming small errors in determining the true rotation offsets between state estimate $\*s_t$ and the true state $\*s^*_t$, we consider the random variable $R'\tilde{R}_t^\top\*t^i$ where $R'$ represents the random rotation matrix corresponding to small angular deviations}~\cite{barfoot_pose_2011}\chl{. Using $R'\tilde{R}_t^\top\*t^i$, we approximate the covariance matrix $\Sigma^{(i)}_t$ as}
 \begin{align*}
	 \cmhl{\Sigma^{(i)}_t} & \cmhl{\approx \Sigma^i_t + \E[(R'-I)(\tilde{R}_t^\top\*t^i)(\tilde{R}_t^\top\*t^i)^\top(R'-I)^\top]}
	 \\
	 \cmhl{[ \Sigma^{(i)}_t]_{i'j'}} 
	 &\cmhl{\approx [ \Sigma^i_t ]_{i'j'} + \E[(\*r'_{i'})^\top (\tilde{R}_t^\top\*t^i)(\tilde{R}_t^\top\*t^i)^\top (\*r'_{j'})]}
	 \\
	 &\cmhl{ = [ \Sigma^i_t ]_{i'j'} + \mathrm{Tr}\left((\tilde{R}_t^\top\*t^i)(\tilde{R}_t^\top\*t^i)^\top \E[(\*r'_{i'})(\*r'_{j'})^\top] \right)}
	 \\
	 &\cmhl{ = [ \Sigma^i_t ]_{i'j'} + \mathrm{Tr}\left((\tilde{R}_t^\top\*t^i)(\tilde{R}_t^\top\*t^i)^\top Q_{i'j'} \right),} \numberthis
 \end{align*}
 \chl{where $(\*r'_{i})^\top$ represents the $i$th row vector in $R'-I$. Since errors in $\tilde{R}$ depend on the DNN output, we specify $R'$ through the empirical distribution of the angular deviations in $\tilde{R}$ as observed for the trained DNN on the training and validation data, and precompute the expectation $Q_{i'j'}$ for each $(i', j')$ pair.}     

The samples of state estimate position error $\{\Delta \*x_t^{(1)}, \ldots, \Delta \*x_t^{(N_C)}\}$ represent both inaccuracy in the DNN estimation as well as uncertainties due to environmental factors. If the DNN approximation fails at the input corresponding to the state estimate $\*s_t$, the estimated position errors at candidate states would lead to a wide range of different values for the state estimate position errors. Similarly, if the environment map $\mathcal{M}$ near the state estimate $\*s_t$ contains repetitive features, the position errors computed from candidate states would be different and hence indicate high uncertainty.

%% file: outlier_wt.tex
\subsection{Outlier Weights}
Since the candidate states $\{\*s^{1}_t \ldots, \*s^{N_C}_t\}$ are selected randomly, some position error samples may correspond to the local depth map and camera image pairs for which the DNN performs poorly. Thus, we compute outlier weights $\{\*w^{(1)}_t, \ldots, \*w^{(N_C)}_t\}$ corresponding to the position error samples $\{\Delta \*x_t^{(1)}, \ldots, \Delta \*x_t^{(N_C)}\}$ to mitigate the effect of these erroneous position error values in determining the protection levels. \ghl{We compute outlier weights in each of the $x, y,$ and $z$-dimensions separately, since the DNN approximation might not necessarily fail in all of its outputs. An example of this scenario is when the input camera image and local map contain features such as building edges that can be used to robustly determine errors along certain directions but not others.}

For computing the outlier weights $\*w_t^{(i)} = [w^{(i)}_{x, t}, w^{(i)}_{y ,t}, w^{(i)}_{z, t} ]$ associated with the $i$th position error value $\Delta \*x_t^{(i)} = [\Delta x_t^{(i)}, \Delta y_t^{(i)}, \Delta z_t^{(i)}]$, we employ the robust Z-score based outlier detection technique~\cite{iglewicz_how_1993}. The robust Z-score is used in a variety of anomaly detection approaches due to its resilience to outliers~\cite{rousseeuw_anomaly_2018}. We apply the following operations in each dimension $X=x, y,$ and $z$:
\begin{enumerate}
	\item We compute the Median Absolute Deviation statistic~\cite{iglewicz_how_1993} ${M\negthinspace AD}_X$ using all position error values $\{\Delta X_t^{(1)}, \ldots, \Delta X_t^{(N_C)}\}$ 
	\begin{equation}
		{M\negthinspace AD}_X = \median(|\Delta X_t^{(i)} - \median(\Delta X_t^{(i)})|).
	\end{equation}
	\item Using the statistic ${M\negthinspace AD}_X$, we compute the robust Z-score $\mathcal{Z}^{(i)}_X$ for each position error value $\Delta X_t^{(i)}$
	\begin{equation}
		\mathcal{Z}^{(i)}_X = \frac{|\Delta X_t^{(i)} - \median(\Delta X_t^{(i)})|}{{M\negthinspace AD}_X}.
	\end{equation}
	The robust Z-score $\mathcal{Z}^{(i)}_X$ is high if the position error $\Delta \*x^{(i)}$ deviates from the median error with a large value when compared with the median deviation value.
	\item We compute the outlier weights $\{w^{(1)}_X, \ldots, w^{(N_C)}_X\}$ from the robust Z-scores $\{\mathcal{Z}^{(1)}_X, \ldots, \mathcal{Z}^{(N_C)}_X\}$ by applying the softmax operation~\cite{goodfellow_deep_2016} such that the sum of weights is unity
	\begin{equation}
		w^{(i)}_{X, t} = \frac{e^{-\gamma \cdot \mathcal{Z}^{(i)}_X}}{\sum_{j=1}^{N_C}e^{-\gamma \cdot \mathcal{Z}^{(j)}_X}},
	\end{equation}
	where $\gamma$ denotes the scaling coefficient in the softmax function. We set $\gamma=0.6745$ as the approximate inverse of the standard normal distribution evaluated at $3/4$ to make the scaling in the statistic consistent with the standard deviation of a normal distribution \cite{iglewicz_how_1993}. A small value of outlier weight $w^{(i)}_{X, t}$ indicates that the position error $\Delta X_t^{(i)}$ is an outlier.  
\end{enumerate}
\ghl{For brevity, we extract the diagonal variances associated with each dimension for all position error samples}
\begin{align*}
	\gmhl{(\sigma^2_{x, t})^{(i)}} &\gmhl{= [\Sigma^{(i)}_t]_{11}}\\
	\gmhl{(\sigma^2_{y, t})^{(i)}} &\gmhl{= [\Sigma^{(i)}_t]_{22}}\\
	\gmhl{(\sigma^2_{z, t})^{(i)}} &\gmhl{= [\Sigma^{(i)}_t]_{33}}. \numberthis 
\end{align*}


%% file: prob_dist.tex
\subsection{Probability Distribution of Position Error}
\label{sec:prob_dist}
\ghl{We construct a probability distribution in each of the $X = x, y$ and $z$-dimensions from the previously obtained samples of position errors $\Delta X^{(i)}_t$, variances $(\sigma^2_{X, t})^{(i)}$ and outlier weights $w^{(i)}_{X, t}$.} We model the probability distribution using the Gaussian Mixture Model (GMM) distribution~\cite{lindsay_mixture_1995}
\begin{align*}
	\P(\rho_{X, t}) &= \sum_{i=1}^{N_C} w^{(i)}_{X, t} \mathcal{N}\left(\Delta X_t^{(i)}, (\sigma^2_{X, t})^{(i)}\right), \numberthis
\end{align*}
where 
\begin{enumerate}
	\item[--] $\rho_{X, t}$ denotes the position error random variable
	\item[--] $\mathcal{N}(\mu, \sigma^2)$ is the Gaussian distribution with mean $\mu$ and variance $\sigma^2$.
\end{enumerate}

\ghl{The probability distributions $\P(\rho_{x, t})$, $\P(\rho_{y, t})$ and $\P(\rho_{z, t})$ incorporate both aleatoric uncertainty from the DNN-based covariance and epistemic uncertainty from the multiple DNN evaluations associated with different candidate states.} \chl{Both the position error and covariance matrix depend on the rotation error point estimate from CMRNet for transforming the error values to the vehicle reference frame. Since each DNN evaluation for a candidate state estimates the rotation error independently, the epistemic uncertainty incorporates the effects of errors in DNN-based estimation of both rotation and translation.} The epistemic uncertainty is reflected in the multiple GMM components and their weight coefficients, which represent the different possible position error values that may arise from the same camera image measurement and the environment map. The aleatoric uncertainty is present as the variance in each possible value of the position error represented by the individual components.  

%% file: pl_comp.tex
\subsection{Protection Levels}
\ghl{We compute the protection levels along the lateral, longitudinal and vertical directions using the probability distributions obtained in the previous section. Since the position errors are in the vehicle reference frame, the $x,y$ and $z$-dimensions coincide with the lateral, longitudinal and the vertical directions, respectively.} First, we obtain the cumulative distribution function $\textrm{CDF}(\cdot)$ for each probability distribution
\begin{align*}
	\gmhl{\textrm{CDF}(\rho_{X, t})}&\gmhl{= \sum_{i=1}^{N_C} w^{(i)}_{X, t} \Phi\left(\frac{\rho_{X, t}-\Delta X_t^{(i)}}{(\sigma_{X, t})^{(i)}}\right)} \numberthis
\end{align*}
where $\Phi(\cdot)$ is the cumulative distribution function of the standard normal distribution.
\ghl{Then, for a specified value of the integrity risk $IR$, we compute the protection level $PL$ in lateral, longitudinal and vertical directions from equation}~\ref{eqn:pl} \ghl{using the $\textrm{CDF}$ as the probability distribution. For numerical optimization, we employ a simple interval halving method for line search or the bisection method}~\cite{burden_numerical_2011}. 
\ghl{To account for both positive and negative errors, we perform the optimization both using $\textrm{CDF}$ (supremum) and $1-\textrm{CDF}$ (infemum) with $IR/2$ as the integrity risk and use the maximum absolute value as the protection level.}

The computed protection levels consider heavy-tails in the GMM probability distribution of the position error that arise because of the different possible values of the position error that can be computed from the available camera measurements and environment map. Our method computes large protection levels when many different values of position error may be equally probable from the measurements, resulting in larger tail probabilities in the GMM, and small protection levels only if the uncertainty from both aleatoric and epistemic sources is small.

%% file: results.tex
\section{Experimental Results}

\subsection{Real-World Driving Dataset}
We use the KITTI visual odometry dataset~\cite{geiger_are_2012} to evaluate the performance of the protection levels computed by our approach. The dataset was recorded around Karlsruhe, Germany over multiple driving sequences and contains images recorded by multiple on-board cameras, along with ground truth positions and orientations. Additionally, the dataset contains LiDAR point cloud measurements which we use to generate the environment map corresponding to each sequence. Since our approach for computing protection levels just requires a monocular camera sensor, we use the images recorded by the left RGB camera in our experiments. We use the sequences $00$, $03$, $05$, $06$, $07$, $08$ and $09$ from the dataset based on the availability of a LiDAR environment map. We use sequence $00$ for validation of our approach and the rest of the sequences are utilized in training our DNN. The experimental parameters are provided in Table \ref{tab:param}. 

\subsection{LiDAR Environment Map}
\begin{figure}
    \centering
    \begin{floatrow}
    \capbtabbox{%
    \begin{tabular}{@{}ll@{}}
        \toprule
        Parameter & Value \\
        \midrule
		Integrity risk $IR$ & $0.01$\\
		Candidate state maximum translation offset $t_{max}$ & $1.0$ m \\
		Candidate state maximum rotation offset $r_{max}$ & $5^{\circ}$ \\
        Number of candidate states $N_C$ & $24$\\
        \ghl{Lateral alarm limit $AL_{lat}$} & $0.85$ m \\
		\ghl{Longitudinal alarm limit $AL_{lon}$} & $1.50$ m \\
		\ghl{Vertical alarm limit $AL_{vert}$} & $1.47$ m \\ 
        \bottomrule

    \end{tabular}
    }{%
      \caption{Experimental parameters}%
      \label{tab:param}%
    }
    \ffigbox{%
      \includegraphics[width=0.45\textwidth]{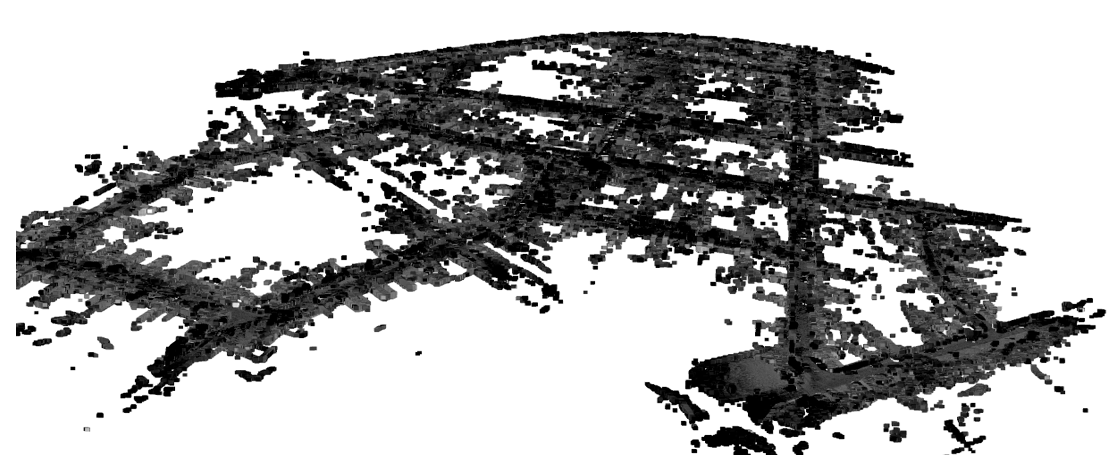}%
    }{%
      \caption{3D LiDAR environment map from KITTI dataset sequence 00~\cite{geiger_are_2012}.}%
      \label{fig:kitti_data}
    }
    \end{floatrow}
\end{figure}

To construct a precise LiDAR point cloud map $\mathcal{M}$ of the environment, we exploit the openly available position and orientation values for the dataset computed via Simultaneous Localization and Mapping~\cite{caselitz_monocular_2016}. Similar to~\cite{cattaneo_cmrnet_2019}, we aggregate the LiDAR point clouds across all time instances. Then, we detect and remove sparse outliers within the aggregated point cloud by computing Z-score~\cite{iglewicz_how_1993} of each point in a $0.1$ m local neighborhood. We discarded the points which had a higher Z-score than $3$. Finally, the remaining points are down sampled into a voxel map of the environment $\mathcal{M}$ with resolution of $0.1$ m. The corresponding map for sequence 00 in the KITTI dataset is shown in Fig. \ref{fig:kitti_data}. For storing large maps, we divide the LiDAR point cloud sequences into multiple overlapping parts and construct separate maps of roughly $500$ Megabytes each.    

\subsection{DNN Training and Testing Datasets}
We generate the training dataset for our DNN in two steps. First, we randomly select a state estimate $s_t$ at time $t$ from within a $2$ m translation and a $10^\circ$ rotation of the ground truth positions and orientations in each driving sequence. The translation and rotation used for generating the state estimate is utilized as the ground truth position error $\Delta \*x^*_t$ and orientation error $\Delta \*r^*_t$. Then, using the LiDAR map $\mathcal{M}$, we generate the local depth map $L(\*s_t, \mathcal{M})$ corresponding to the state estimate $\*s_t$ and use it as the DNN input along with the camera image $I_t$ from the driving sequence data. The training dataset comprises of camera images from $11455$ different time instances, with the state estimate selected at runtime so as to have different state estimates for the same camera images in different epochs.  

Similar to the data augmentation techniques described in~\cite{cattaneo_cmrnet_2019}, we 
\begin{enumerate}
	\item Randomly change contrast, saturation and brightness of images,
	\item Apply random rotations in the range of $\pm 5^\circ$ to both the camera images and local depth maps,
	\item Horizontally mirror the camera image and compute the local depth map using a modified camera projection matrix. 
\end{enumerate}   
All three of these data augmentation techniques are used in training CMRNet in the first half of the optimization process. However, for training CovarianceNet, we skip the contrast, saturation and brightness changes during the second half of the optimization so that the DNN can learn real-world noise features from camera images. 

We generate the validation and test datasets from sequence $00$ in the KITTI odometry dataset, which is not used for training. We follow a similar procedure as the one for generating the training dataset, except we do not augment the data. The validation dataset comprises of randomly selected $100$ time instances from sequence $00$, while the test dataset contains the remaining $4441$ time instances in sequence $00$.

\subsection{Training Procedure}
 
We train the DNN using stochastic gradient descent. Directly optimizing via the maximum likelihood loss term $\mathcal{L}_{\textrm{MLE}}(\cdot)$ might suffer from instability caused by the interdependence between the translation error $\Delta \tilde{\*x}$ and covariance $\tilde{\Sigma}$ outputs~\cite{skafte_reliable_2019}. Therefore, we employ the mean-variance split training strategy proposed in~\cite{skafte_reliable_2019}: First, we set $(\alpha_{\textrm{Huber}}=1, \alpha_{\textrm{MLE}}=1, \alpha_{\textrm{Ang}}=1)$ and only optimize the parameters of CMRNet till validation error stops decreasing. Next, we set $(\alpha_{\textrm{Huber}}=0, \alpha_{\textrm{MLE}}=1, \alpha_{\textrm{Ang}}=0)$ and optimize the parameters of CovarianceNet. We alternate between these two steps till validation loss stops decreasing. Our DNN is implemented using the PyTorch library~\cite{paszke_pytorch_2019} and takes advantage of the open-source implementation available for CMRNet~\cite{cattaneo_cmrnet_2019} as well as the available pretrained weights for initialization. Similar to CMRNet, all the layers in our DNN use the leaky RELU activation function with a negative slope of $0.1$. We train the DNN on using a single NVIDIA Tesla P40 GPU with a batch size of $24$ and learning rate of $10^{-5}$ selected via grid search.

\subsection{Metrics}
We evaluate the lateral, longitudinal and vertical protection levels computed using our approach using the following three metrics (with subscript $t$ dropped for brevity):
\begin{enumerate}
	\item \ghl{\textbf{Bound gap} measures the difference between the computed protection levels $PL_{lat}, PL_{lon}, PL_{vert}$ and the true position error magnitude during nominal operations (protection level is less than the alarm limit and greater than the position error)}
	\begin{align*}
		\gmhl{BG_{lat}} &\gmhl{= \textrm{avg}(PL_{lat} - |\Delta x^*|)} \\
		\gmhl{BG_{lon}} &\gmhl{= \textrm{avg}(PL_{lon} - |\Delta y^*|)} \\
		BG_{vert} &= \textrm{avg}(PL_{vert} - |\Delta z^*|), \numberthis
	\end{align*}
	where
	\begin{enumerate}
		\item[--] $BG_{lat}, BG_{lon}$ and $BG_{vert}$ denote bound gaps in lateral, longitudinal and vertical dimensions respectively
		\item[--] \ghl{$\textrm{avg}(\cdot)$ denotes the average computed over the test dataset for which the value of protection level is greater than position error and less than the alarm limit}
	\end{enumerate}
	\ghl{A small bound gap value $BG_{lat}, BG_{lon}, BG_{vert}$ is desirable since it implies that the algorithm both estimates the position error magnitude during nominal operations accurately and has low uncertainty in the prediction. We only consider the bound gap for nominal operations, since the estimated position is declared unsafe when the protection level exceeds the alarm limit.}
	\item \textbf{Failure rate} measures the total fraction of time instances in the test data sequence for which the computed protection levels $PL_{lat}, PL_{lon}, PL_{vert}$ are smaller than the true position error magnitude
	\begin{align*}
		\gmhl{FR_{lat}} &\gmhl{= \frac{1}{T_{\textrm{max}}}\sum_{t=1}^{T_{\textrm{max}}}\mathbbm{1}_t\left(PL_{lat} < |\Delta x^*|\right)} \\
		\gmhl{FR_{lon}} &\gmhl{= \frac{1}{T_{\textrm{max}}}\sum_{t=1}^{T_{\textrm{max}}}\mathbbm{1}_t\left(PL_{lon} < |\Delta y^*|\right)}\\
		\gmhl{FR_{vert}} &\gmhl{= \frac{1}{T_{\textrm{max}}}\sum_{t=1}^{T_{\textrm{max}}}\mathbbm{1}_t\left(PL_{vert} < |\Delta z^*|\right),} \numberthis
	\end{align*}
	where
	\begin{enumerate}
		\item[--] $FR_{lat}, FR_{lon}$ and $FR_{vert}$ denote failure rates for lateral, longitudinal and vertical protection levels, respectively
		\item[--] $\mathbbm{1}_t(\cdot)$ denotes the indicator function computed using the protection level and true position error values at time $t$. The indicator function evaluates to $1$ if the event in its argument holds true, and otherwise evaluates to $0$
		\item[--] $T_{\textrm{max}}$ denotes the total time duration of the test sequence  
	\end{enumerate}
	The failure rate $FR_{lat}, FR_{lon}, FR_{vert}$ should be consistent with the specified value of the integrity risk $IR$ to meet the safety requirements.
	\item \ghl{\textbf{False alarm rate} is computed for a specified alarm limit $AL_{lat}, AL_{lon},  AL_{vert}$ in the lateral, longitudinal and vertical directions and measures the fraction of time instances in the test data sequence for which the computed protection levels $PL_{lat}, PL_{lon},  PL_{vert}$ exceed the alarm limit $AL_{lat}, AL_{lon},  AL_{vert}$ while the position error magnitude is within the alarm limits.} We first define the following integrity events
	\begin{align*}
		\gmhl{\Omega_{lat, PL}} &\gmhl{= (PL_{lat} > AL_{lat}) }\\
		\gmhl{\Omega_{lat, PE}} &\gmhl{= (|\Delta x^*| > AL_{lat}) }\\
		\gmhl{\Omega_{lon, PL}} &\gmhl{= (PL_{lon} > AL_{lon}) }\\
		\gmhl{\Omega_{lon, PE}} &\gmhl{= (|\Delta y^*| > AL_{lon}) }\\
		\gmhl{\Omega_{vert, PL}} &\gmhl{= (PL_{vert} > AL_{vert}) }\\
		\gmhl{\Omega_{vert, PE}} &\gmhl{= (|\Delta z^*| > AL_{vert}).} \numberthis
	\end{align*}
	The complement of each event is denoted by $\bar{\Omega}$. Next, we define the counts for false alarms $N_{X, FA}$, true alarms $N_{X, TA}$ and the number of times the position error exceeds the alarm limit $N_{X, PE}$ with $X=lat, lon$ and $vert$ 
	\begin{align*}
		\gmhl{N_{X, FA}} &\gmhl{= \sum_{t=1}^{T_{\textrm{max}}}\mathbbm{1}_t\left(\Omega_{X, PL} \cap \bar{\Omega}_{X, PE}\right)} \\
		\gmhl{N_{X, TA}} &\gmhl{= \sum_{t=1}^{T_{\textrm{max}}}\mathbbm{1}_t\left(\Omega_{X, PL} \cap \Omega_{X, PE}\right)} \\
		\gmhl{N_{X, PE}} &\gmhl{= \sum_{t=1}^{T_{\textrm{max}}}\mathbbm{1}_t\left(\Omega_{X, PE}\right).} \numberthis
	\end{align*}  
	Finally, we compute	the false alarm rates $FAR_{lat}, FAR_{lon}, FAR_{vert}$ after normalizing the total number of position error magnitudes lying above and below the alarm limit $AL$  
	\begin{align*} 
		\gmhl{FAR_{X}} &\gmhl{= \frac{N_{X, FA} \cdot ({T_{\textrm{max}}} - N_{X, PE})}{N_{X, FA} \cdot ({T_{\textrm{max}}} - N_{X, PE}) + N_{X, TA} \cdot N_{X, PE}}.} \numberthis
	\end{align*}
\end{enumerate}

\subsection{Results}
\begin{figure}[t!]
	\centering
	\includegraphics[width=\textwidth]{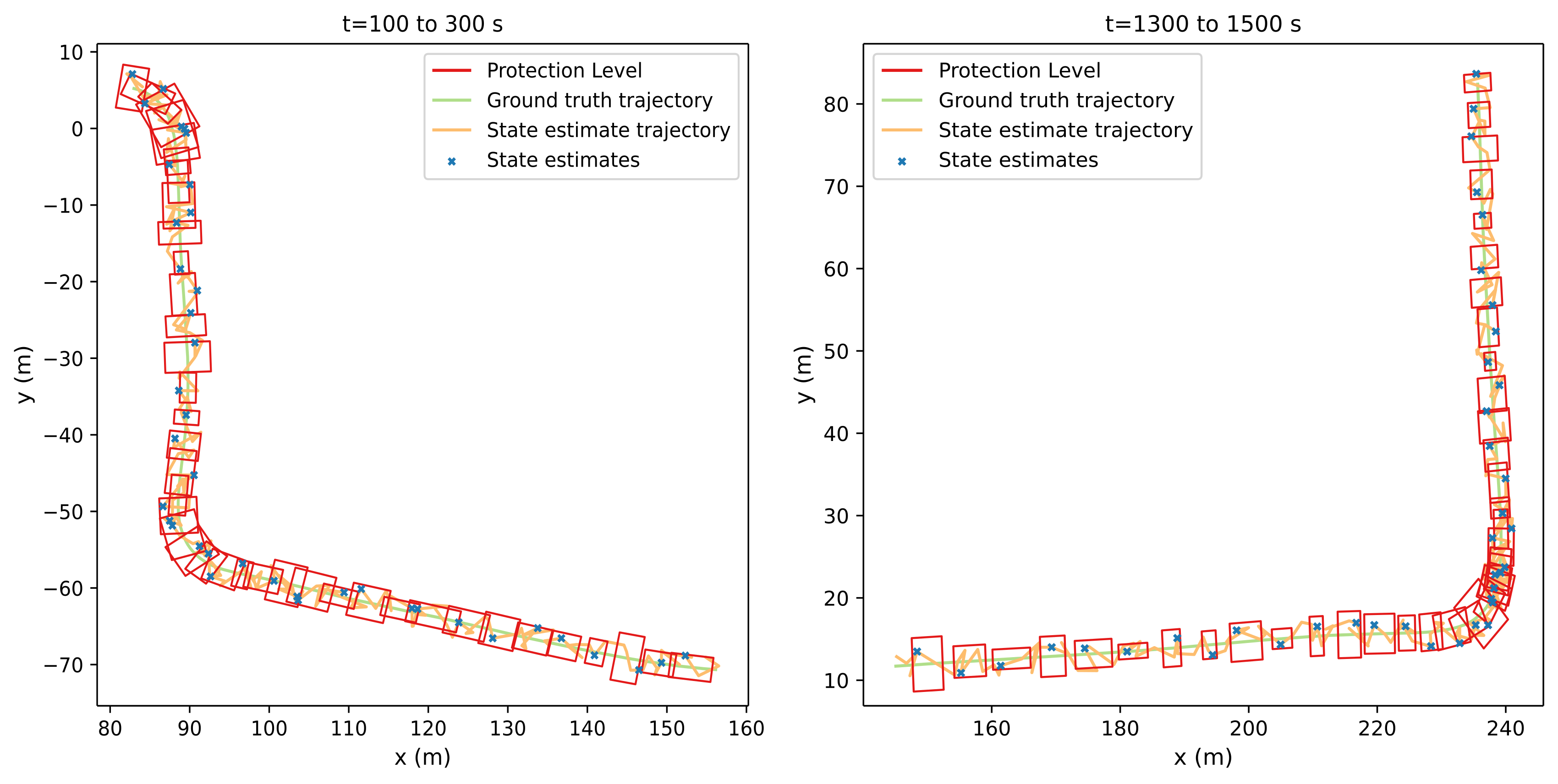}
	\caption{\ghl{Lateral and longitudinal protection level results on the test sequence in real-world dataset. We show protection levels for two subsets of the total sequence, computed at every $5$ s intervals. The protection levels successfully enclose the state estimates in $\sim 99\%$ of the cases.}}
	\label{fig:qual_hpl}
\end{figure}

\begin{figure}[t!]
	\centering
	\includegraphics[width=\textwidth]{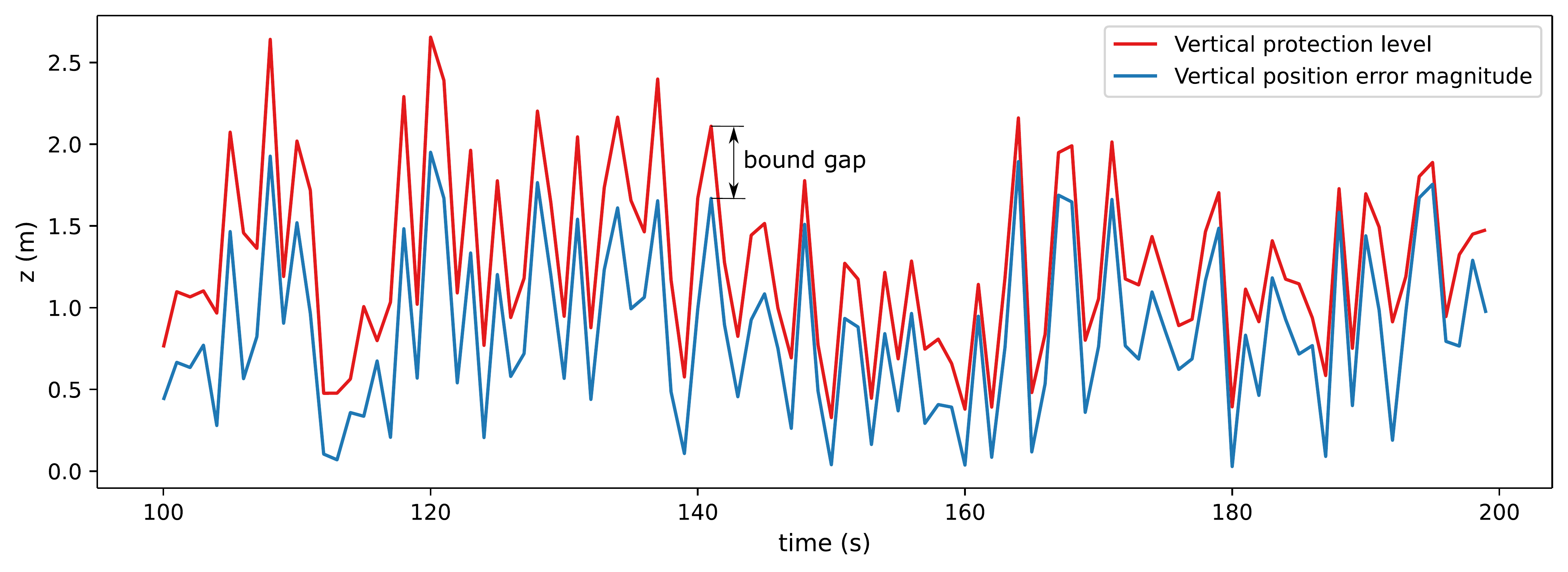}
	\caption{\ghl{Vertical protection level results on the test sequence in real-world dataset. We show protection levels for a subset of the total sequence. The protection levels successfully enclose the position error magnitudes with a small bound gap.}}
	\label{fig:qual_vpl}
\end{figure}

Fig.~\ref{fig:qual_hpl} \ghl{shows the lateral and longitudinal protection levels computed by our approach on two $200$ s subsets of the test sequence.} For clarity, protection levels are computed at every $5$th time instance. Similarly, Fig. \ref{fig:qual_vpl} shows the vertical protection levels along with the vertical position error magnitude in a subset of the test sequence. As can be seen from both the figures, the computed protection levels successfully enclose the position error magnitudes at a majority of the points ($\sim 99\%$) in the visualized subsequences. Furthermore, the vertical protection levels are observed to be visually closer to the position error as compared to the lateral and longitudinal protection levels. This is due to the superior performance of the DNN in determining position errors along the vertical dimension, which is easier to learn since all the camera images in the dataset are captured by a ground-based vehicle.

\begin{figure}
	\centering
	\includegraphics[width=0.9\linewidth]{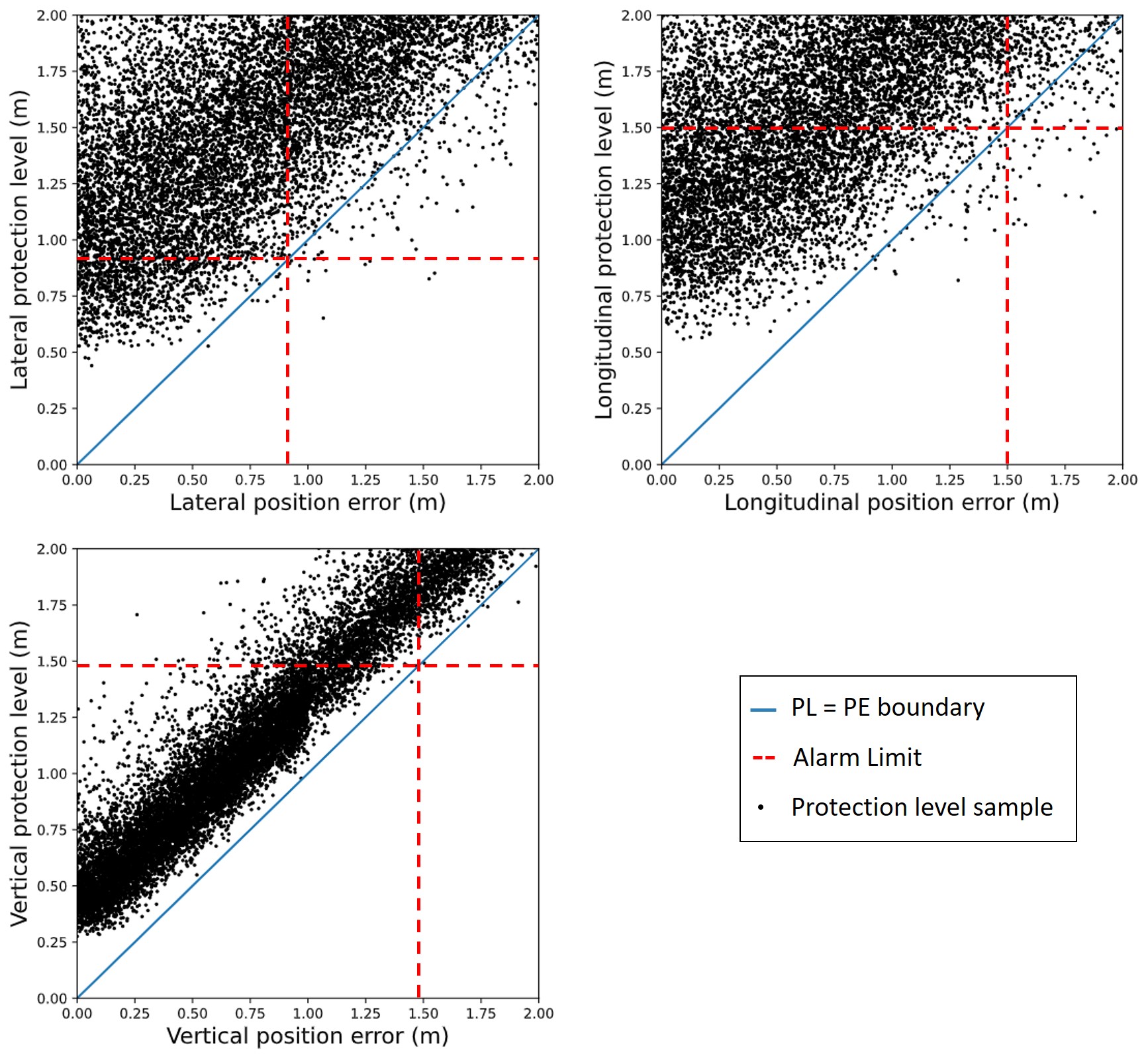}  
	\caption{\ghl{Integrity diagram results for the lateral, longitudinal and vertical protection levels. The diagram contains protection levels evaluated across $15000$ different state estimates and camera images randomly selected from the test sequence. A majority of the samples are close to and greater than the position error magnitude, validating the applicability of the computed protection levels as a robust safety measure.}}
	\label{fig:integ_plot}
\end{figure}

Fig.~\ref{fig:integ_plot} displays the integrity diagrams generated after the Stanford-ESA integrity diagram proposed for SBAS integrity~\cite{tossaint_stanford_2007}. \ghl{The diagram is generated from $15000$ samples of protection levels corresponding to randomly selected state estimates and camera images within the test sequence.}\chl{ For protection levels each direction, we set the alarm limit} (Table~\ref{tab:param}) \chl{based on the specifications suggested for mid-size vehicles in}~\cite{reid_localization_2019}, \chl{beyond which the state estimate is declared unsafe to use.} \ghl{The lateral, longitudinal and vertical protection levels are greater than the position error magnitudes in $\sim 99$\% cases, which is consistent with the specified integrity requirement. Furthermore, a large fraction of the failures is in the region where the protection level is greater than the alarm limit and thus the system has been correctly identified to be under unsafe operation.}    

We conducted an ablation study to numerically evaluate the impact of our proposed epistemic uncertainty measure and outlier weighting method in computing protection levels. We evaluated protection levels in three different cases: Incorporating DNN covariance, epistemic uncertainty and outlier weighting (\textsc{VAR+EO}); incorporating just the DNN covariance and epistemic uncertainty with equal weights assigned to all position error samples (\textsc{VAR+E}); and only 
using the DNN covariance (\textsc{VAR}). For \textsc{VAR}, we constructed a Gaussian distribution using the DNN position error output and diagonal variance entries in each dimension. Then, we computed protection levels from the inverse cumulative distribution function of the Gaussian distribution corresponding to the specified value of integrity risk $IR$. Table \ref{tab:met} summarizes our results. \ghl{Incorporating the epistemic uncertainty in computing protection levels improved the failure rate from $0.05$ in lateral protection levels, $0.05$ in longitudinal protection levels and $0.03$ in vertical protection levels to within $0.01$ in all cases. This is because the covariance estimate from the DNN provides an overconfident measure of uncertainty, which is corrected by our epistemic uncertainty measure. Furthermore, incorporating outlier weighting reduced the average nominal bound gap by about $0.02$ m in lateral protection levels, $0.05$ m in longitudinal protection levels, and $0.05$ m in vertical protection levels as well as false alarm rate by about $0.02$ for each direction while keeping the failure rate within the specified integrity risk requirement.}

\ghl{The mean bound gap between the lateral protection levels computed from our approach and the position error magnitudes in the nominal cases is smaller than a quarter of the width of a standard U.S. lane. In the longitudinal direction, the bound gap is somewhat larger since fewer visual features are present along the road for determining the position error using the DNN. The corresponding value in the vertical dimension is smaller, owing to the DNN's superior performance in determining position errors and uncertainty in the vertical dimension. This demonstrates the applicability of our approach to urban roads.}

For an integrity risk requirement of $0.01$, the protection levels computed by our method demonstrate a failure rate equal to or within $0.01$ as well. However, further lowering the integrity risk requirement during our experiments either did not similarly improve the failure rate or caused a significant increase in the bound gaps and the false alarm rate. A possible reason is that the uncertainty approximated by our approach through both the aleatoric and epistemic measures fails to act as an accurate uncertainty representation for smaller integrity risk requirements than $0.01$. Future research would consider more and varied training data, better strategies for selecting candidate states, and different DNN architectures to meet smaller integrity risk requirements.

\chl{A shortcoming of our approach is the large false alarm rate exhibited by the computed protection levels in Table}~\ref{tab:met}\chl{. The large value results both from the inherent noise in the DNN-based estimation of position and rotation error as well as from frequently selecting candidate states that result in large outlier error values. A future work direction for reducing the false alarm rate is to explore strategies for selecting candidate states and mitigating outliers.}

\chl{A key advantage offered by our approach is its application to scenarios where a direct analysis of the error sources in the state estimation algorithm is difficult, such as when feature rich visual information is processed by a machine learning algorithm for estimating the state. In such scenarios, our approach computes protection levels separately from the state estimation algorithm by both evaluating a data-driven model of the position error uncertainty and characterizing the epistemic uncertainty in the model outputs.}

\begin{table*}\centering
	\renewcommand{\arraystretch}{1.1}
	\begin{tabular}{@{}rrrrrrrrrrrr@{}} \toprule
		& \multicolumn{3}{c}{Lateral PL} & \phantom{abc} & \multicolumn{3}{c}{Longitudinal PL} & \phantom{abc} & \multicolumn{3}{c}{Vertical PL} \\
		\cmidrule{2-4} \cmidrule{6-8} \cmidrule{10-12}
		&  $BG$(m) & $FR$ &  $FAR$ && $BG$(m) & $FR$ &  $FAR$ && $BG$(m) & $FR$ &  $FAR$\\ \midrule
		\textsc{VAR+EO} & $0.49$ & $0.01$ & $0.47$ 
					   && $0.77$ & $0.01$ & $0.40$ 
		               && $0.38$ & $< \negmedspace 0.01$ & $0.14$\\
		\textsc{VAR+E} & $0.51$ & $0.01$ & $0.49$ 
		              && $0.82$ & $0.01$ & $0.43$
					  && $0.43$ & $< \negmedspace 0.01$ & $0.16$\\
		\textsc{VAR} & $0.42$ & $0.05$ & $0.45$ 
				    && $0.64$ & $0.05$ & $0.36$
					&& $0.30$ & $0.02$ & $0.12$\\ \bottomrule
	\end{tabular}
	\caption{\ghl{Evaluation of lateral, longitudinal and vertical protection levels from our approach. We compare protection levels computed by our trained model using DNN covariance, epistemic uncertainty and outlier weighting (VAR+EO), DNN covariance and epistemic uncertainty (VAR+E) and only DNN covariance (VAR). Incorporating epistemic uncertainty results in lower failure rate, while incorporating outlier weights reduces bound gap and false alarm rate.}}
	\label{tab:met}
\end{table*}

%% file: conclusion.tex
\section{Conclusions}

In this work, \ghl{we presented a data-driven approach for computing lateral, longitudinal and vertical protection levels associated with a given state estimate from camera images and a 3D LiDAR map of the environment.} Our approach estimates both aleatoric and epistemic measures of uncertainty for computing protection levels, thereby providing robust measures of localization safety. We demonstrated the efficacy of our method on real-world data in terms of bound gap, failure rate and false alarm rate. \ghl{Results show that the lateral, longitudinal and vertical protection levels computed from our method enclose the position error magnitudes with $0.01$ probability of failure and less than $1$ m bound gap in all directions, which demonstrates that our approach is applicable to GNSS-denied urban environments.}
